\documentclass{article}

\usepackage{arxiv}

\usepackage{booktabs} 
\usepackage[usestackEOL]{stackengine}
\strutlongstacks{T}

\usepackage{tikz}
\usetikzlibrary{positioning,shapes,shadows,arrows}
\usepackage{xcolor}
\usepackage{amsmath}
\usepackage{amssymb}
\usepackage{cuted}

\DeclareMathOperator*{\argmin}{arg\,min} 

\title{Interplanetary Transfers via Deep Representations of the Optimal Policy and/or of the Value Function}

\author{
  Dario Izzo\thanks{Dario Izzo is the corresponding author}, Ekin \"{O}zt\"{u}rk, Marcus M\"{a}rtens \\
  Advanced Concepts Team \\
  European Space Agency \\
  Noordwijk, 2201 AZ \\
  Netherlands \\
  \texttt{dario.izzo@esa.int} \\
  \texttt{ekin.ozturk@esa.int} \\
  \texttt{marcus.maertens@esa.int}
}

\begin{document}
\maketitle
\begin{abstract}
A number of applications have been recently proposed based on deep networks and interplanetary trajectories. These approaches often rely on the availability of a large number of optimal trajectories to learn from. In this paper we introduce a new method to quickly create millions of optimal spacecraft trajectories from a single nominal trajectory.
Apart from the generation of the nominal trajectory, no additional optimal control problems need to be solved as all the trajectories, by construction, satisfy Pontryagin's minimum principle and the relevant transversality conditions.
We then consider deep feed forward neural networks and benchmark three learning methods on the created dataset: policy imitation, value function learning and value function gradient learning. 
Our results are shown for the case of the interplanetary trajectory optimization problem of reaching Venus orbit, with the nominal trajectory starting from the Earth. 
We find that both policy imitation and value function gradient learning are able to learn the optimal state feedback, while in the case of value function learning the optimal policy is not captured, only the final value of the optimal propellant mass is. 
\end{abstract}


\maketitle

\section{Introduction}
\label{sec:intro}
The use of deep neural networks (DNNs) for the guidance navigation and control of space systems (spacecraft, landers, etc..) is a prolific area of research as witnessed by the increasing number of results that appeared recently on these topics \cite{sanchez2018real, sprague2019learning, cheng2019real, tailor2019learning, furfaro2018deep, izzo2018survey, zhu2019fast}. Most of this body of work can be divided into DNNs approximating the optimal policy $\mathbf u$ (e.g. the optimal thrust profile), and DNNs approximating the value function $v$ (e.g. optimal propellant consumption). The final goal of this line of research is to develop DNNs that can be part of the on-board software to steer the spacecraft and substitute the guidance and control loops currently used. Furthermore, accurate predictions of important quantities (e.g. propellant mass) for optimal transfers can provide advantages during preliminary design phases of trajectories where many options are typically screened and evaluated.

While the deployment of DNNs for trajectory design is promising, it has been shown over a large range of different application domains that the performance of DNNs depends heavily on the availability of suitable datasets for training. While most research literature is concerned with improving network architectures (number of layers, regularization, activation functions, etc.) or training procedures, in practice, it is often simply the lack of data that puts a limit on what any DNN architecture is capable of achieving. 

The computation of a large-scale dataset of trajectorie, including the optimal control profile and the value function is challenging, as a huge number of optimal control problems need to be solved. Since numerical instabilities and convergence issues are dominant, solution procedures are generally hard to automate and demanding in terms of computational resources.

The contribution of this paper is two-fold: First, we show how to create a large dataset of optimal trajectories without the need to solve more than one optimal control problem. This decreases the dataset generation time, in comparison to previous methods, by orders of magnitude. Second, we use the dataset to learn several control policies for an interplanetary mission to reach the orbit of Venus. 

Building on the work of~\cite{sanchez2018real, tailor2019learning, sanchez2016learning}, we propose three different learning tasks, which we call \emph{policy imitation}, \emph{value function learning} and \emph{value function gradients learning}. We are able to approximate, in the former case, the optimal policy (i.e. the optimal control profile) and, in the latter two cases, the value function (i.e. the solution to the corresponding Hamilton Jacobi Bellman equation). In the last case we find that the learned value function can also be satisfactorily used to compute the optimal policy.

This work is structured as follows: Section~\ref{sec:background} provides the details of the nominal transfer to Venus designed as a mass-optimal trajectory by solving the two point boundary value problem resulting from Pontryagin's minimum principle. Based on this nominal trajectory, Section~\ref{sec:building} describes the new method for the generation of a large dataset of different trajectories to the Venus orbit whose mass-optimality is still ensured by Pontryagin's minimum principle. Section~\ref{sec:experiments} describes different strategies for the training of DNNs on such a dataset. The resulting controllers are evaluated experimentally in Section~\ref{sec:results} and we conclude in Section~\ref{sec:conclusion}.

\section{Background}
\label{sec:background}
We consider, in the International Celestial Reference Frame (ICRF) heliocentric frame, the motion of a spacecraft of mass $m$ whose velocity and position we indicate with $\mathbf r$ and $\mathbf v$. The spacecraft is equipped with an ion thruster having specific impulse $I_{sp}$ and capable of delivering a maximum thrust $c_1$ regardless of the spacecraft distance from the Sun. 
We describe the spacecraft state via its mass and the modified equinoctial elements $\mathbf x = [p, f, g, h, k, L]^T$ as originally defined by Walker et al.~\cite{walker}.

The set of differential equations describing the spacecraft motion are then:

\begin{equation}
\begin{array}{l}
\dot p = \sqrt{\frac p\mu} \frac{2p}{w} f_t \\
\dot f = \sqrt{\frac p\mu} \left\{ f_r\sin L + \left[ (1+w)\cos L + f \right] \frac{f_t}w - (h\sin L-k\cos L)\frac{g\cdot f_n}{w} \right\} \\
\dot g = \sqrt{\frac p\mu} \left\{ - f_r\cos L + \left[ (1+w)\sin L + g \right] \frac{f_t}w + (h\sin L-k\cos L)\frac{f\cdot f_n}{w} \right\} \\
\dot h = \sqrt{\frac p\mu} \frac{s^2f_n}{2w}\cos L \\
\dot k = \sqrt{\frac p\mu} \frac{s^2f_n}{2w}\sin L \\
\dot L = \sqrt{\frac p\mu}\left\{\mu\left(\frac wp\right)^2 + \frac 1w\left(h\sin L-k\cos L\right) f_n\right\} \\
\dot m = - \frac{\sqrt{f_r^2+f_t^2+f_n^2}}{I_{sp}g_0}
\end{array}
\end{equation}
\normalsize
where, $w = 1 + f\cos L + g\sin L$ and $s^2 = 1 + h^2 + k^2$ and $f_r, f_t, f_n$ are the radial, tangential and normal components of the force generated by the spacecraft propulsion system. The gravity parameter is denoted with $\mu$ and the gravitational acceleration at sea level with $g_0$.

It is useful to rewrite the equations above using a more concise notation. We thus introduce the matrices $\mathbf B$ and $\mathbf D$ defined as:
\begin{equation}
\sqrt{\frac \mu p} \mathbf B(\mathbf x) = \left[
\begin{array}{ccc}
0 & \frac {2p}w & 0 \\
 \sin L & [(1+w)\cos L + f]\frac 1w  & - \frac gw (h\sin L-k\cos L)  \\
- \cos L & [(1+w)\sin L + g]\frac 1w  & \frac fw (h\sin L-k\cos L) \\
0 & 0  & \frac 1w \frac{s^2}{2}\cos L \\
0 & 0  & \frac 1w \frac{s^2}{2}\sin L \\
0 & 0  & \frac 1w (h\sin L - k\cos L) \\
\end{array}
\right]
\end{equation}
and 
\begin{equation}
\mathbf D(\mathbf x) = 
\left[
\begin{array}{cccccc}
0 & 0 & 0 & 0 & 0 & \sqrt{\mu p}\left(\frac{w}{p}\right)^2
\end{array}
\right]
\end{equation}
and rewrite the equations of motion in the form:
\begin{equation}
\left\{
\begin{array}{l}
\dot {\mathbf x} =  \frac {c_1 u(t)} m \mathbf B(\mathbf x)  \mathbf{\hat i_\tau}  + \mathbf D(\mathbf x) \\
\dot m = -c_2 u(t)
\end{array}
\right.
\label{eq:eom}
\end{equation}
where the spacecraft thrust is now indicated by $c_1 u \mathbf{\hat i_\tau} = [f_r, f_t, f_n]^T$ and bounded by $|u(t)| \le 1$ and $|\mathbf{\hat i_\tau}(t)| = 1$. We will use the notation $\mathbf u\in \mathcal U$ to indicate the feasible control space. Furthermore, we define $c_2 = 1 / (I_{sp}\ g_0)$. Note that in order to steer the spacecraft, at each instant we control the throttle magnitude $u(t) \in [0,1]$ and the thrust direction $\mathbf{\hat i_\tau}$. We refer to these control variables also with the single symbol $\mathbf u = [u, \mathbf{\hat i_\tau}]$. 

\subsection{The Low-Thrust Problem}
\label{sec:problem}
We consider here a free time orbital transfer problem, that is finding the controls $u(t)$ and $\mathbf{\hat i_\tau}(t)$ defined in $[0, t_f]$ and the transfer time $t_f$ so that the functional:
\begin{equation}
\label{eq:J}
J(u(t), t_f) = \int_{0}^{t_f} \left\{ u - \epsilon \log{[u(1-u)]} \right\} dt
\end{equation}
is minimized, and the spacecraft is steered from its initial mass $m_0$ and some initial point $\mathbf x_0$ to some final mass $m_f$ and some final point $\mathbf x_f \in \mathcal S_f \subset \mathbb R^6$. The functional $J$, following the work of Bertrand and Epenoy \cite{bertrand2002new}, is parameterised by a continuation parameter $\epsilon \in [0,1]$ which activates a logarithmic barrier smoothing the problem and ensuring that the constraint $u(t) \in [0,1]$ is always satisfied. Clearly for $\epsilon \rightarrow 0$ the functional becomes $J = (m_0 - m_f) / c_2$, equivalent to minimizing the propellant mass.

\subsection{Consequences of Pontryagin's Minimum Principle}
\label{sec:pontryagin}
Following the work of Pontryagin~\cite{pontryagin}, we can infer the necessary conditions for an optimal solution of this problem by applying Pontryagin's minimum principle.\footnote{Note that we have stated a minimization problem, hence the conditions are actually slightly different from the ones originally derived in Pontryagin's work.}


Let us introduce the co-states $\boldsymbol\lambda, \lambda_m$ as continuous functions defined in $[0, t_f]$ and define the Hamiltonian:
\begin{equation}
\label{eq:hamil}
    \mathcal H(\mathbf x, m, \boldsymbol\lambda, \lambda_m, \mathbf u)  = \frac {c_1 u} m \boldsymbol\lambda^T \mathbf B(\mathbf x) \mathbf{\hat i_\tau}  +  \lambda_L \sqrt{\frac \mu {p^3}}w^2 - c_2 \lambda_m u + \left\{ u - \epsilon \log[u(1-u)] \right\}
\end{equation}

and the system of equations:

\begin{equation}
\label{eq:eom_costates}
\left\{
\begin{array}{l}
\dot {\mathbf x} =  \frac{\partial \mathcal H}{ \boldsymbol\lambda}=  \frac {c_1 u(t)} m \mathbf B \mathbf{\hat i_\tau}(t) + \mathbf D  \\
\dot m =  \frac{\partial \mathcal H}{\partial \lambda_m} = -c_2 u(t)  \\
\dot{ \boldsymbol\lambda} = - \frac{\partial \mathcal H}{\partial \mathbf x} \\
\dot{\lambda}_m = -\frac{\partial \mathcal H}{\partial m} \\
\end{array}
\right.
\end{equation}

The explicit form of the various derivatives appearing in the equations above is reported in Appendix~\ref{app}.

Along an optimal trajectory, the Hamiltonian must be zero (free terminal time problem) and minimal w.r.t. the choices of $u$ and $\mathbf{\hat i_\tau}$. For the optimal thrust direction $\mathbf{\hat i_\tau}^*$ it follows that
\begin{equation}
\label{eq:optimal_direction}
\mathbf{\hat i_\tau}  = \mathbf{\hat i_\tau}^*(t) = - \frac{ \mathbf B^T\boldsymbol\lambda}{| \mathbf B^T\boldsymbol\lambda|}
\end{equation}
where the time dependence on the right hand side follows both from the co-states and from $\mathbf B$ and has not been indicated explicitly for brevity. While for the optimal throttle $u^*$, necessarily:
\begin{equation}
\label{eq:optimal_magnitude}
u(t) = u^*(t) = \frac{2\epsilon}{2\epsilon + SF(t) + \sqrt{4\epsilon^2 + SF(t)^2}}
\end{equation}
where we introduce a switching function
\begin{equation}
\label{eq:switching}
SF(t) = 1 - \frac{c_1}{m }| \mathbf B^T\boldsymbol\lambda|-c_2\lambda_m.
\end{equation}

\subsubsection{The two point boundary value problem}
The test case considered in this work is that of a transfer from any $\mathbf x_0, m_0$ to Venus' orbit (not a rendezvous). In this case the final value for the mass and the final value for the anomaly $L$ are left free so that some transversality conditions apply  ($\lambda_{L_f} = 0, \lambda_{m_f} = 0$). For any starting state $\mathbf x_0$ and $m_0$, we need to choose the values for $\boldsymbol \lambda_0$ and $\lambda_{m_0}$ and the time $t_f$ such that solving the initial value problem for Eq.(\ref{eq:eom_costates}) results in a final state that matches the arrival conditions at Venus orbit, the transversality conditions and the free-time condition on the Hamiltonian. Formally, we introduce the shooting function:
\begin{equation}
\label{eq:shooting}
\phi(\boldsymbol \lambda_0, \lambda_{m_0}, t_f) =  [p_f-p_V,f_f-f_V,g_f-g_V,h_f-h_V,k_f-k_V, \lambda_{L_f}, \lambda_{m_f}, \mathcal H_f]
\end{equation}
where we indicate with a subscript $V$ the modified equinoctial elements of Venus orbit and with the subscript $f$ the final values of the modified equinoctial elements resulting from numerically integrating  Eq.(\ref{eq:eom_costates}) for a time $t_f$ and from the initial conditions $\mathbf x_0, m_0, \boldsymbol  \lambda_0, \lambda_{m_0}$. We have thus transformed our optimal control problem into the problem: $\phi(\boldsymbol \lambda_0, \lambda_{m_0}, t_f)  = 0$.

\begin{figure}[tb]
    \centering
    \includegraphics[width=0.7\columnwidth]{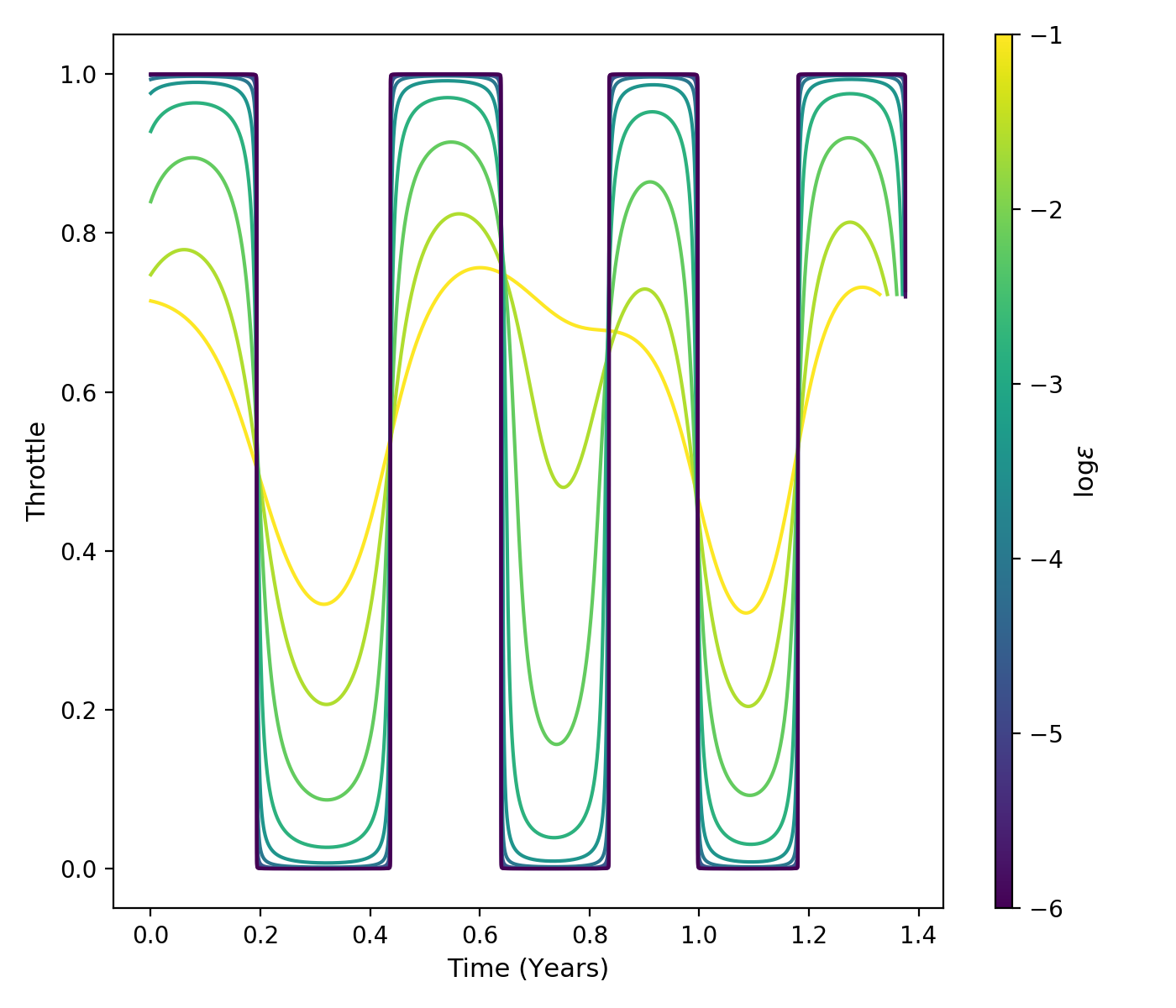}
    \caption{Solution of the two point boundary value problem for the throttle magnitude \emph{u(t)} and decreasing values of $\epsilon$.
    \label{fig:homotopy}}
\end{figure}

\begin{figure}[tb]
    \centering
    \includegraphics[width=0.6\columnwidth]{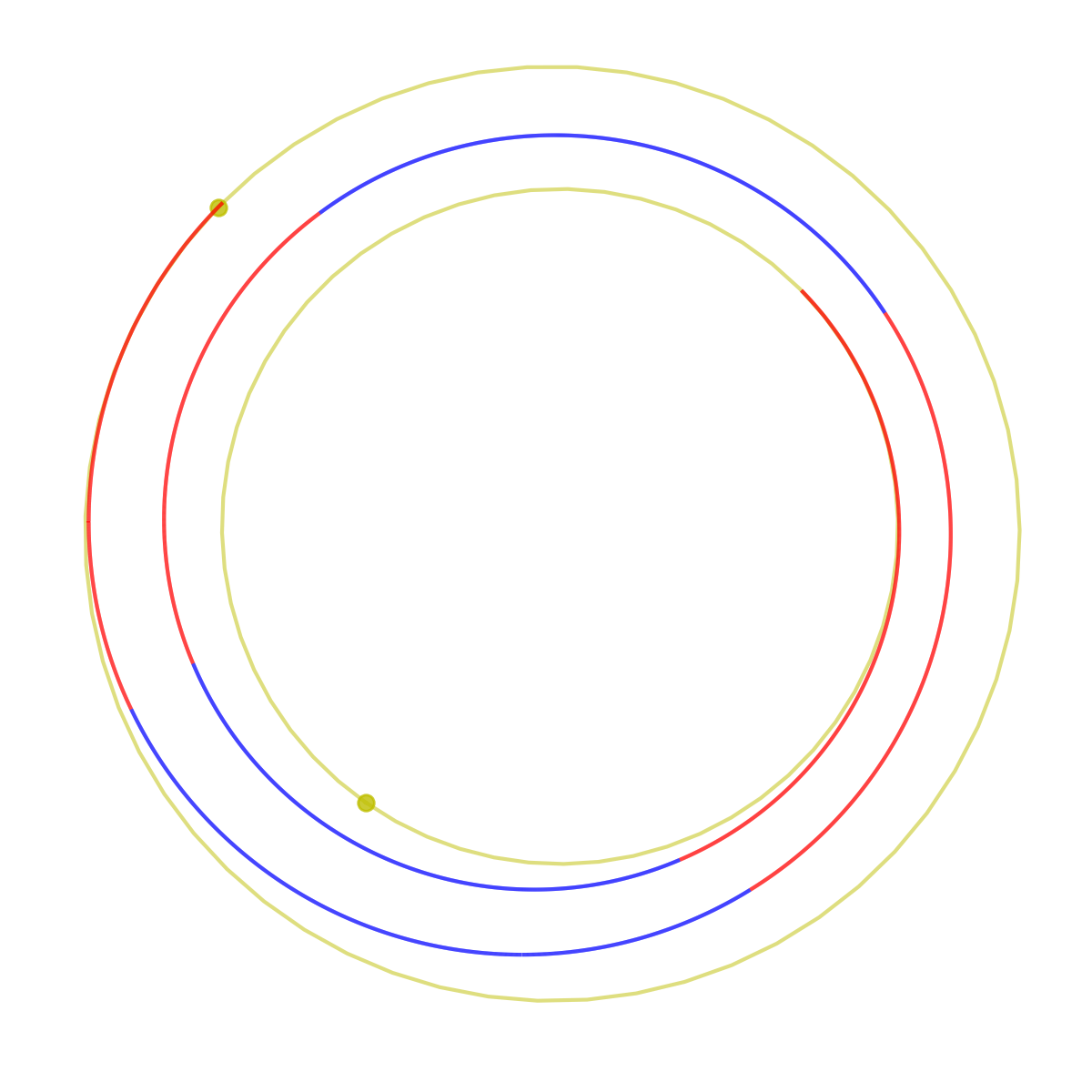}
    \caption{Nominal trajectory to Venus orbit. Thrust arcs are indicated in red.
    \label{fig:nominal_trajectory}}
\end{figure}

\subsubsection{The nominal trajectory}
We consider a spacecraft with mass $m_0 = 1500$ [kg], a nuclear electric propulsion system specified by $I_{sp} = 3800$ [sec] and $c_1 = 0.3$ [N]. We look for the optimal transfer from the Earth to Venus orbit with a launch on the 7th of May 2005. 
The planet ephemerides are computed using JPL low-precision ephemerides. We solve the optimal control problem by solving the problem $\phi(\boldsymbol \lambda_0, \lambda_{m_0}, t_f)  = 0$. Note that this is, essentially, a system of eight equations in eight unknowns and can be solved by root finding methods (e.g. Powell, Levenberg-Marquardt) as well as by SQP or interior point methods (e.g. SNOPT or IPOPT). 
As it is well known (see \cite{haberkorn2004low} for example), the convergence radius for this problem can get rather small, to the point that if we were to try to directly solve the mass optimal problem (i.e. $\epsilon = 0$) we would fail consistently as almost any initial guess on the co-state would not converge. 
However, solving the problem for $\epsilon = 0.1$ is reasonably simple as convergence is frequently achieved when starting with random co-states (we sampled them from a uniform distribution with a standard deviation of 10). Note that we use non dimensional units for the state so that the astronomical unit AU is used for length, the spacecraft initial mass for mass, and the rest is set as to get $\mu = 1.$.
Gradually decreasing $\epsilon$ from $0.1$ down to $10^{-6}$ (as visualized in Figure~\ref{fig:homotopy}), allows us to obtain the final mass optimal trajectory which is visualized in Figure~\ref{fig:nominal_trajectory}. We refer to the final trajectory (with $\epsilon=10^{-6}$) as the \emph{nominal trajectory} in the following. 
The nominal trajectory reaches the orbit of Venus after $t^*_f = 1.376$ [years] and spends $m_p^* = 210.47$ [kg] of propellant.

\subsection{Consequences of Bellman's Principle of Optimality}
\label{sec:bellman}
For $\epsilon = 0$, we can apply Bellman's principle of optimality~\cite{bellman1966dynamic} to the optimal control problem that we have stated in the previous section. We indicate with $v(\mathbf x, m)$ the value function, i.e. the optimal value of the functional defined by Eq.(\ref{eq:J}) for any spacecraft state $\mathbf x, m_0$. Since the value function is time-independent, the Hamilton Jacobi Bellman (HJB) equations can be written as:

\begin{eqnarray}
 0 =  \min_{\mathbf u \in \mathcal U}(u + \nabla v \cdot \mathbf f) \label{eq:HJB1} \\
\mathbf u = \argmin_{\mathbf u \in \mathcal U}(u + \nabla v \cdot \mathbf f)  \label{eq:HJB2}.
\end{eqnarray}
These equations hold in all points where $v(\mathbf x, m)$ is differentiable. We use $\mathbf f$ to denote the right hand side of Eq.(\ref{eq:eom}) including the mass equation. 
Comparing Eq.(\ref{eq:HJB1}) and Eq.(\ref{eq:hamil}), it follows that $\nabla v = [\boldsymbol \lambda, \lambda_m]$ and thus $\mathcal H = u + \nabla v \cdot \mathbf f$. 
Thus, the co-states of Pontryagin's theory are the gradients of the value function introduced in Bellman's theory. 
This fact, albeit rarely exploited in interplanetary trajectory optimization research, provides a convenient basis for the design of new learning procedures, which we introduce and evaluate later in Section~\ref{sec:experiments}.

\section{Building a Dataset of Optimal Trajectories}
\label{sec:building}

\subsection{Reducing the Cost of Computation}
The availability of large-scale datasets is essential to train DNNs. This section shows how to create a large number of optimal trajectories without the need to solve any particular optimal control problem except for the nominal trajectory.
The brute-force approach for computing a dataset of optimal trajectories is to decide on a number of interesting initial states and then solve the corresponding optimal control problem for each( e.g. following all the steps outlined in Section~\ref{sec:background}). This approach scales extremely poorly because of the known computational difficulties associated with solving optimal control problems, both using direct and indirect methods.

Previous work (e.g. Sanchez and Izzo~\cite{sanchez2016learning, sanchez2018real}) deployed a continuation (homotopy) approach to reduce some of the complexity involved by eliminating the need to search for a new initial guess all the time: by perturbing the initial state of a nominal trajectory, the unperturbed co-state provides (most of the time) a good initial guess to solve the newly created two points boundary value problem.
However, assuming that no convergence issues occur, it is still necessary to solve Eq.(\ref{eq:shooting}) for the new initial conditions considered, which incurs a significant computational cost.

In the following, we describe a more efficient way to obtain a similar result avoiding entirely convergence issues, guaranteeing optimality and reducing the computational costs down to the cost of a single numerical propagation of Eq.(\ref{eq:eom_costates}) for a new trajectory. The idea, surprisingly simple and applicable more generally than this Venus orbit acquisition task, is to perform a backward in time propagation of Eq.(\ref{eq:eom_costates}) starting from suitably perturbed final values of the state and co-states of the nominal trajectory. This perturbation needs to be chosen such that the tranversality conditions and the condition on the Hamiltonian are still satisfied.

Formally, consider the nominal optimal trajectory and indicate with $\mathbf x_f, m_f, \boldsymbol \lambda_f, \lambda_{m_f}$ the final values (i.e. at $t^*_f$) of the states and the co-states. Consider the new set of final co-state values:
\begin{equation}
\boldsymbol \lambda'_f = \boldsymbol \lambda_f + \delta \boldsymbol \lambda
\end{equation}

where the perturbation $\delta \boldsymbol \lambda$ is in some ball $B_\rho \in \mathbb R^7$ of size $\rho$ such that the transversality condition on the anomaly ($\delta \lambda_{L} = 0,\ \delta \lambda_{m} = 0$) is satisfied. Note that the trajectory that results from propagating backward in time Eq.(\ref{eq:eom_costates}) from the new final conditions $\mathbf x_f, m_f, \boldsymbol \lambda'_f, \lambda_{m_f}$ is fulfilling all of Pontryagin's necessary conditions for optimality, except $\mathcal H_f = 0$. It is possible to find new final values $m'_f = m_f + \delta m$ and $L'_f = L_f + \delta L$ (e.g. with a simple root finding algorithm) so that the condition on the Hamiltonian is fulfilled. Thus the new trajectory resulting from propagating backward in time Eq.(\ref{eq:eom_costates}) from the conditions $\mathbf x'_f, m'_f, \boldsymbol \lambda'_f, \lambda_{m_f}$ is now fulfilling all of Pontryagin's necessary conditions for optimality. Such a trajectory will neither end where the nominal trajectory ends, nor will it start from where the nominal trajectory starts, but it is nevertheless optimal (with respect to Eq.~\ref{eq:J}) and represents a valid sample to learn from. This procedure reduces the cost of computing one more valid training sample to that of a backward in time integration of Eq.(\ref{eq:eom_costates}) plus the (negligible) cost of a single root finding call to solve $\mathcal H_f(m'_f, L'_f) = 0$.

As we only modify $m_f$ and $L$, our final state is always constrained to be on the orbit of Venus, and since our perturbations for the final co-states are small the timing and direction of thrust maneuvers are close to the nominal trajectory during the backward integration. Together, these imply that it is unlikely, if not impossible, for the final state of the backward integration to go significantly beyond the orbit of Earth during an integration period of $t^*_f=1.376\ [years]$. This is confirmed when looking at the dataset in the next subsection. 

\subsection{The dataset}
\begin{figure}[tb]
    \centering
    \includegraphics[width=0.6\columnwidth]{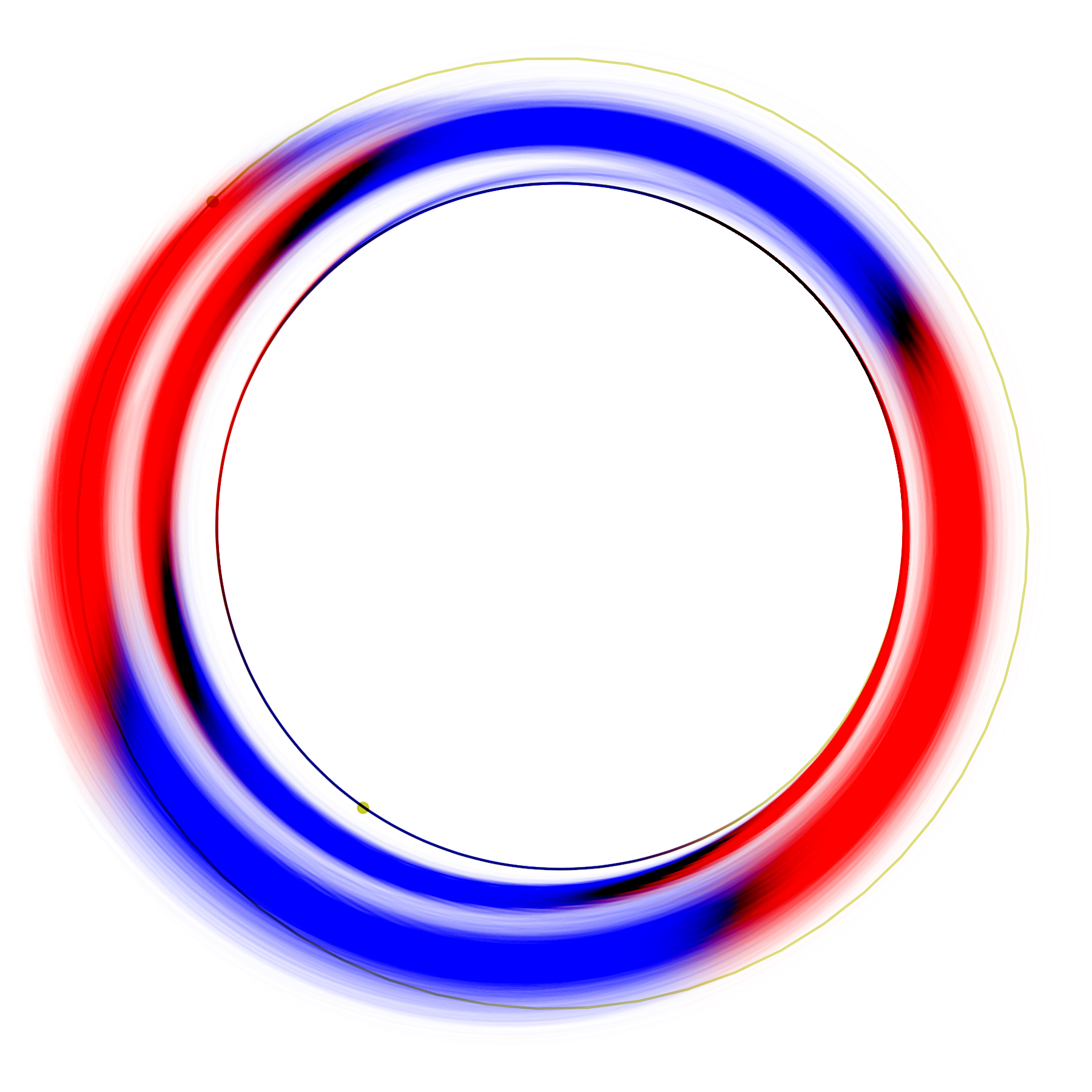}
    \caption{Overlapping optimal dataset trajectories. Thrust arcs are indicated in red. \label{fig:dataset_shadows}}
\end{figure}

To build the particular dataset used in this work, we start from the nominal trajectory of Figure~\ref{fig:homotopy} and generate 500\,000 perturbations of the arrival state $\mathbf x'_f, m'_f, \boldsymbol \lambda'_f$ using $\rho=0.1$ as a perturbation radius. Only perturbations for which the root finder for $\mathcal H_f(L'_f, m'_f)=0$ converged are kept. These valid final states are then integrated backwards in time for $t_f=1.376\ [\textrm{years}]$. At 100 equi-spaced points in time, the state $\mathbf x', m', \boldsymbol \lambda'$ and the optimal policy $u*, \mathbf i_\tau^*$ are sampled and stored in the dataset. Figure~\ref{fig:dataset_shadows} shows an overlapping plot of these trajectories. 

During the dataset generation, 453\,212 feasible perturbations were found. Thus the final dataset consists of 45\,321\,200 samples which took approx. $12$ hours to generate using 20 threads and 25\,000 perturbations per thread on an Intel Xeon E5-2650L v4 processor at 1.70 GHz. Thus, we were able to generate a new optimal trajectory approximately every 1.7 seconds. The bulk of the computational time is spent to integrate Eq.(\ref{eq:eom_costates}) numerically with a relative error tolerance of $10^{-13}$ and an absolute error tolerance of $10^{-13}$ using the LSODA integrator. In Figure 3 we visualize the trajectory dataset in its entirety. Note that all trajectories we will learn from are thus time free, optimal control trajectories.

\section{Experiments}
\label{sec:experiments}
Using the dataset constructed in the previous section, our goal is to evaluate if it is possible to learn a control policy and/or the value function by a deep neural network. 

\subsection{Ideas Behind Our Experiments}
Since this goal can take many forms, we are particularly interested in the following learning tasks:

\begin{figure}[tb]
    \centering
    \includegraphics[width=0.45\columnwidth]{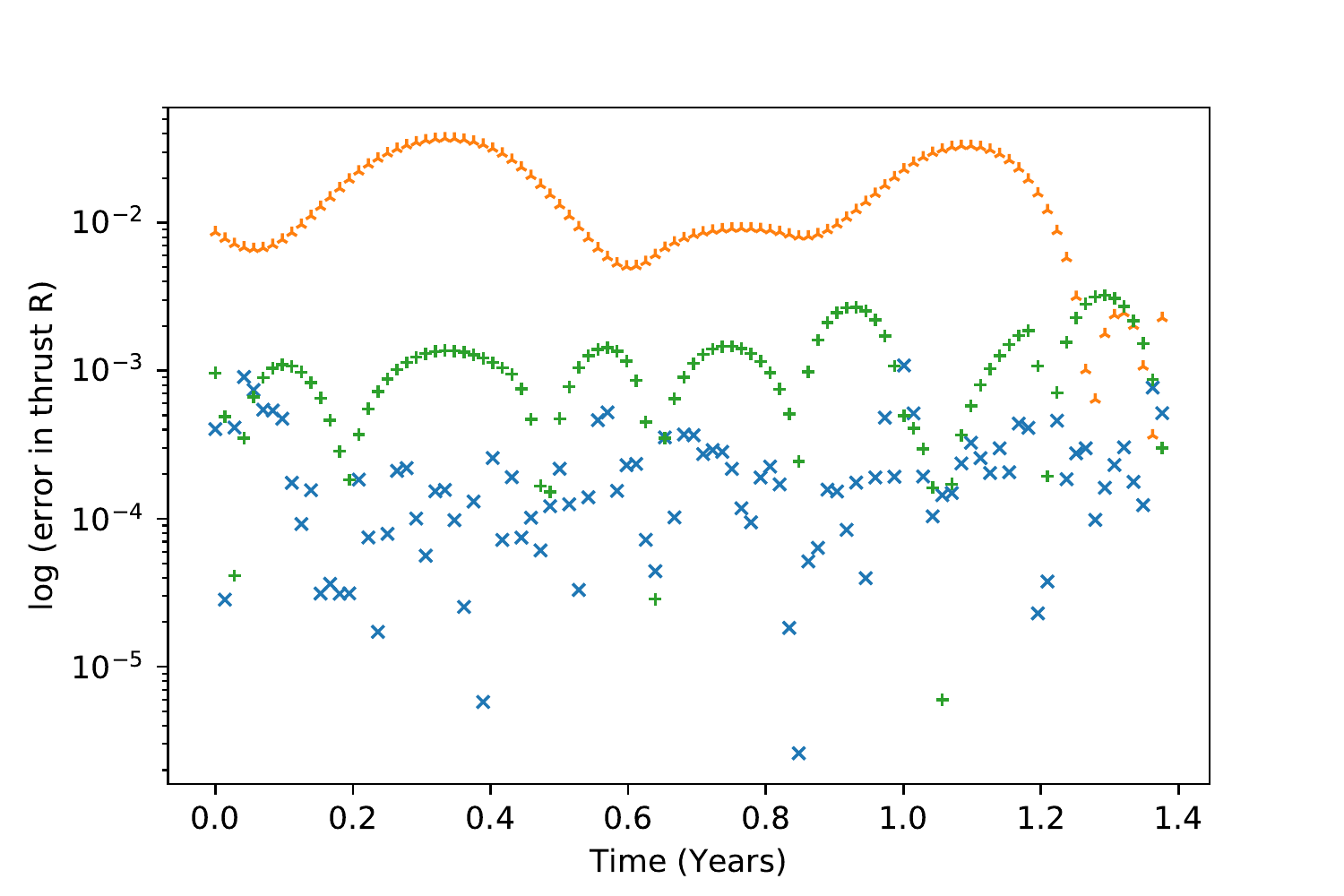}
    \includegraphics[width=0.45\columnwidth]{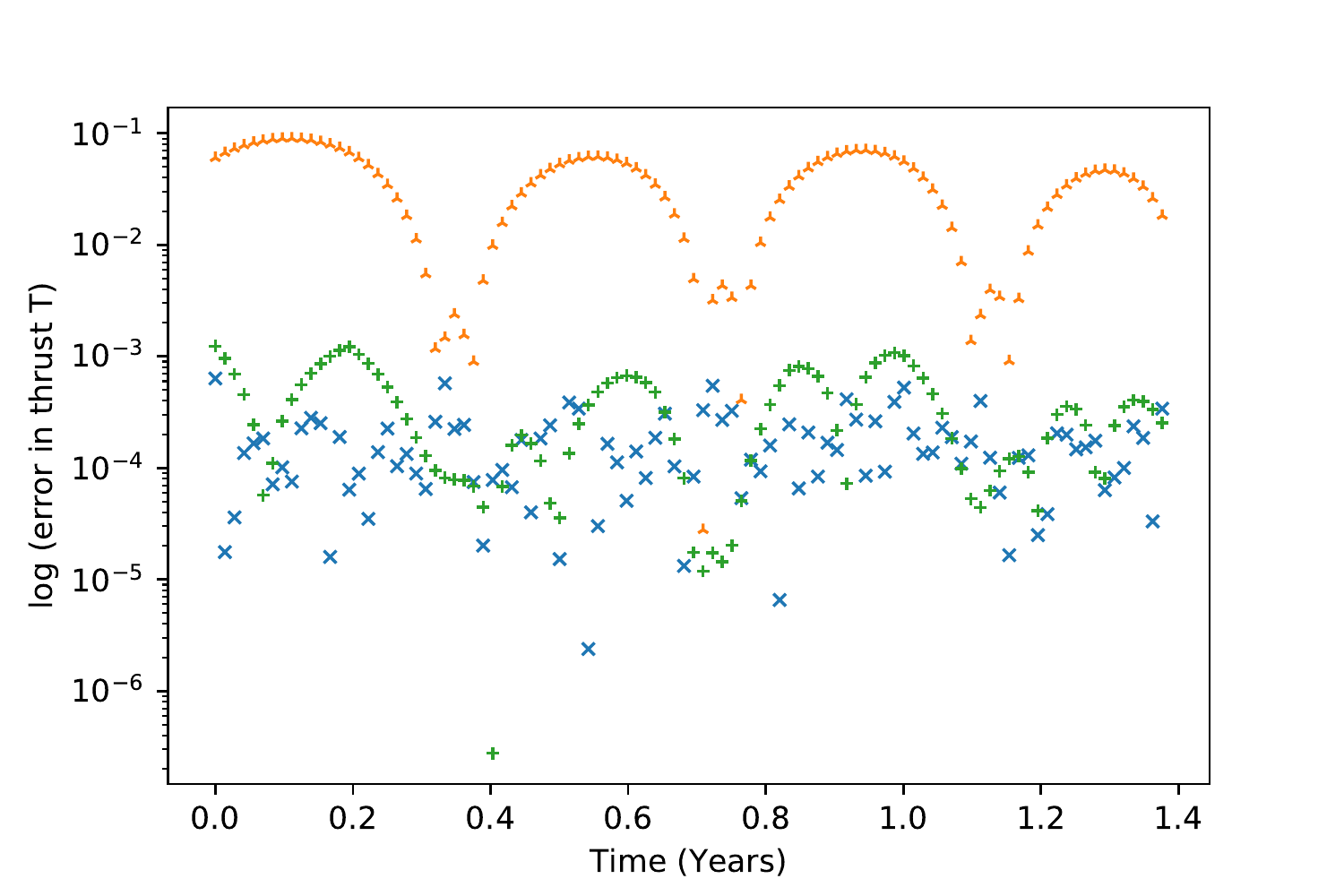}\\
    \includegraphics[width=0.45\columnwidth]{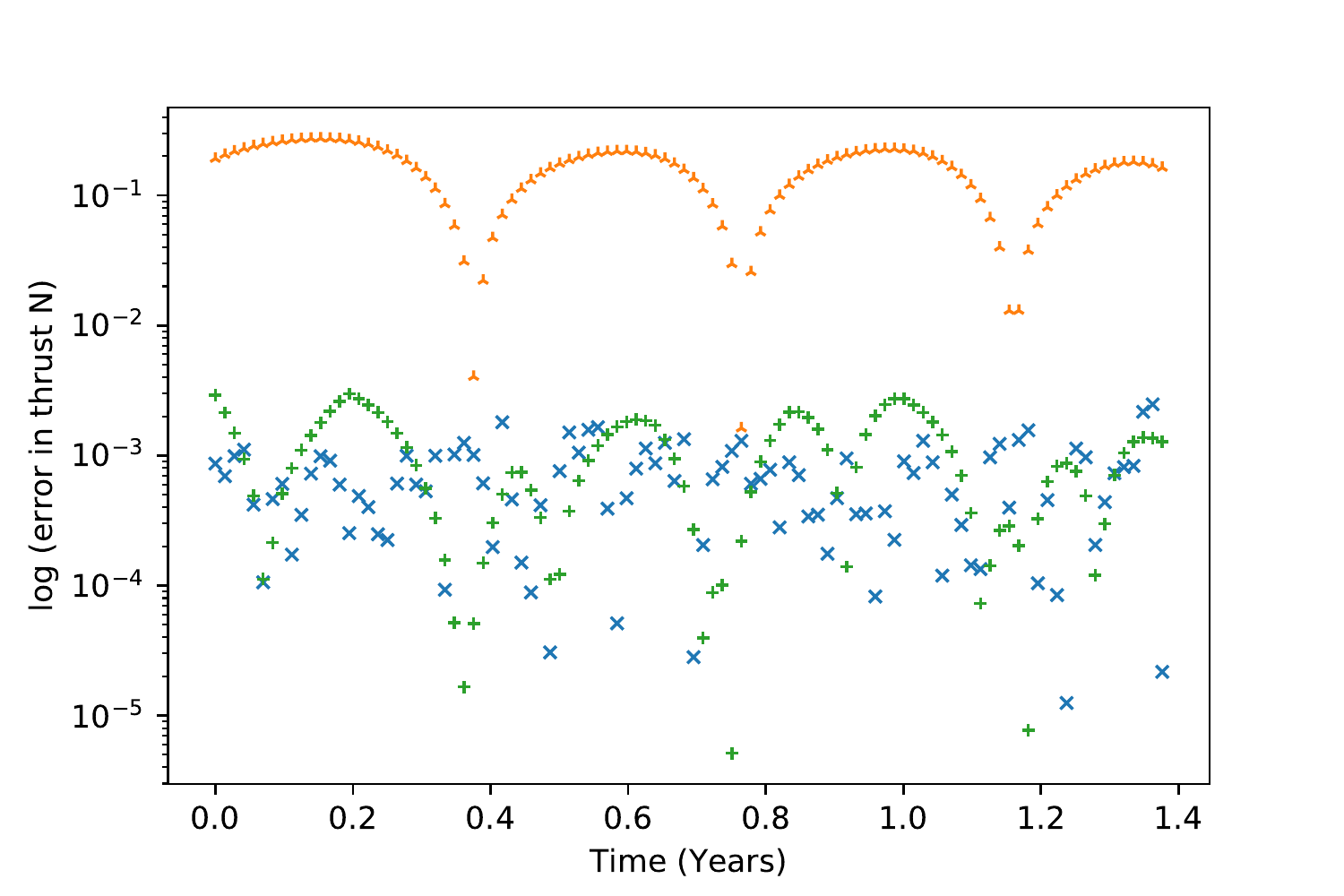}\\
    \includegraphics[width=0.7\columnwidth]{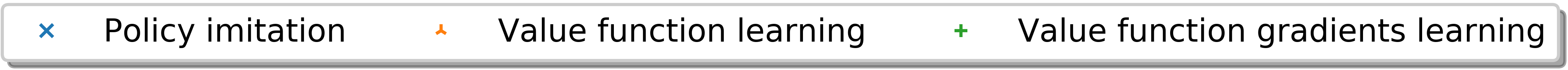}
    \caption{Absolute error for predicting the optimal thrust direction $\mathbf{\hat i_\tau}$  of the nominal trajectory (log-scale).}
    \label{fig:nominal_thrustvector}
\end{figure}

\begin{figure}[tb]
    \centering
    \includegraphics[width=0.9\columnwidth]{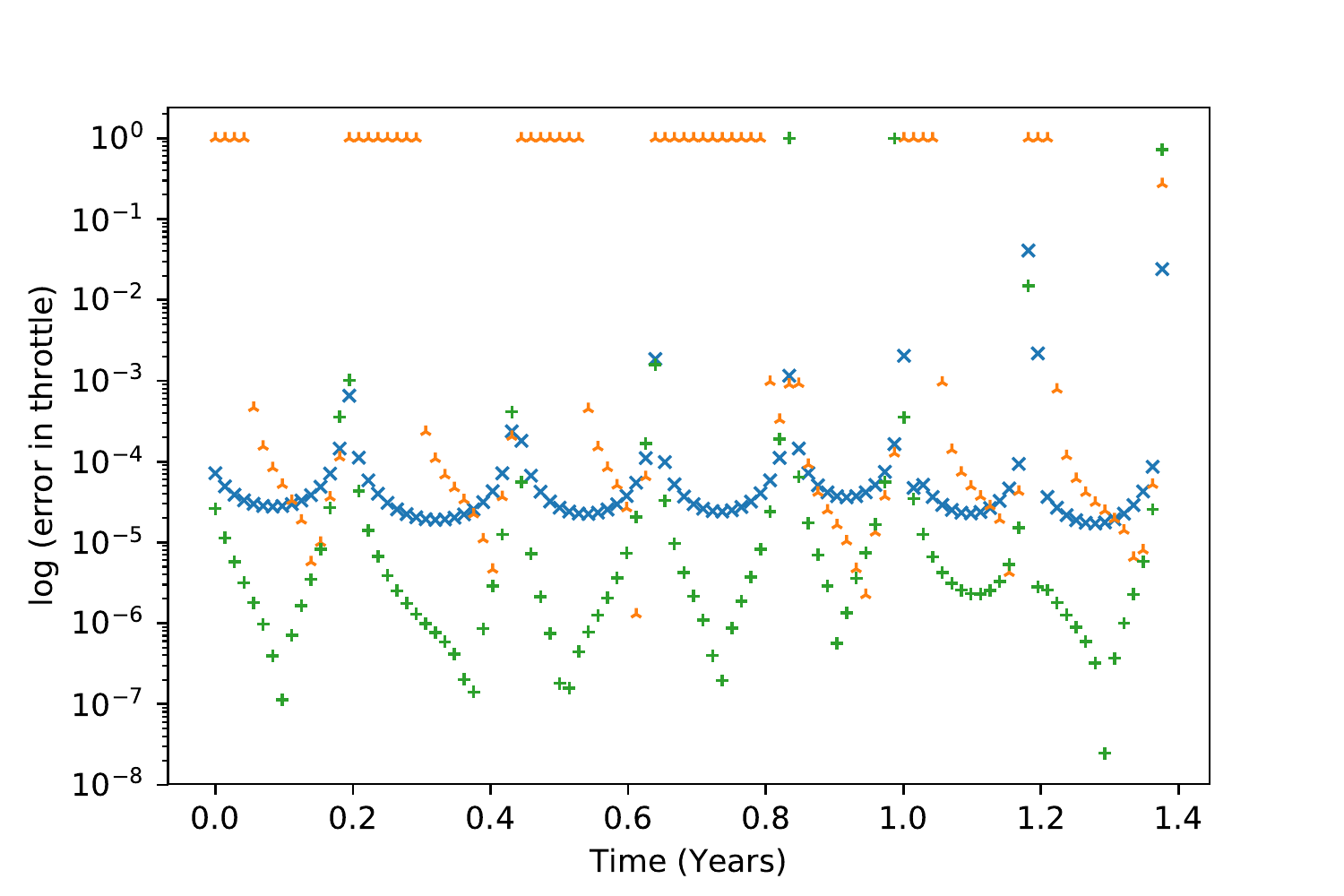}
    \includegraphics[width=0.9\columnwidth]{legend.png}
    \caption{Absolute error for predicting the optimal throttle $u^*$ of the nominal trajectory (log-scale).}
    \label{fig:nominal_throttle}
\end{figure}

\begin{enumerate}
\item \emph{Policy imitation}: In the first experiment we apply the imitation learning pipeline described in \cite{sanchez2018real, izzo2018machine, sprague2019learning} to our training data. The result is a network $\mathcal{N}_{\mathbf u}(\mathbf x, m)$ predicting the optimal $u$ and $\mathbf i_\tau$ directly.

\item \emph{Value function} learning: In the second experiment we train a network $\mathcal{N}_v(\mathbf x, m)$ to predict the value function $v(\mathbf x, m) = m_p / c_2$ (i.e. the optimal propellant mass required to get to Venus' orbit). The loss function used during training and evaluated on a batch of size $b$ is $\ell = \sum_{i=1}^b |v(\mathbf x_i, m_i) - \mathcal{N}_v(\mathbf x_i, m_i)|^2$. Once the network is trained, the Hamilton Jacobi Bellman equation Eq.(\ref{eq:HJB2}) is used to derive the policy $\mathbf u(\mathbf x, m)$. Besides providing a way to compute $\mathbf u(\mathbf x, m)$, this network can also be used for predicting the propellant consumption in preliminary design phases.

\item \emph{Value function gradients} learning: In the third experiment we train a network $\mathcal{N}_{\nabla v}(\mathbf x, m)$ to predict the value function $v(\mathbf x, m)$ and enforce the constraint that $\left[\partial_\mathbf{x}\mathcal{N}_{\nabla v},\partial_m\mathcal{N}_{\nabla v}\right] = \nabla v = \left[\boldsymbol \lambda, \lambda_m\right]$. The idea being that gains are to be achieved, with respect to the previous case, as Eq.(\ref{eq:HJB2}) contains the value function gradient which is here directly learned.  The loss function used during training and evaluated on a batch of size $b$ is thus:
\begin{multline}
\ell = \sum_{i=1}^b \left\{ |v(\mathbf x_i, m_i) - \mathcal{N}_{\nabla v}(\mathbf x_i, m_i)|^2 + \left| \boldsymbol \lambda_i - \frac{\partial \mathcal{N}_{\nabla v}}{\partial \mathbf x}(\mathbf x_i, m_i)\right|^2 + \left| \lambda_{m_i} -  \frac{\partial \mathcal{N}_{\nabla v}}{\partial m}(\mathbf x_i, m_i)\right|^2\right\}
\end{multline}
\end{enumerate}

\subsection{Network Architecture and Training}
The network architectures used for the three experiments are mostly similar except for the depth and the output layers. In all cases, the networks take as input the state $\left[\mathbf x,\ m\right]$ and consist of $n$ hidden layers with softplus activations between each layer (following the results in \cite{tailor2019learning}).

The policy imitation (experiment 1) network consists of $n=4$ hidden layers with 100 units. The output layer has a sigmoid activation and 4 units (throttle and thrust vector).

The value function networks (experiment 2 and 3) consist of $n=9$ hidden layers with 100 units. The output layer has a linear activation and 1 unit. The reason for a linear activation is to avoid normalizing the output, which would also effect the computed gradients used in the loss function. 

The network output can then be differentiated with respect to its inputs to estimate the value function gradients necessary to derive the control policy from Eq.(\ref{eq:HJB2}). In the case of the value function network (experiment 2), these gradients are only used after training to compute the policy. For the value function gradients network (experiment 3), the mean square error of the estimated value function gradients with respect to the co-states is added to the loss-function of the network. Thus the training for the value function gradient network is minimizing both the error of the value-function and its corresponding gradients.

In terms of the training set, the policy network was trained on a normalized dataset where the normalization consists of a standard scaling (i.e. removing the mean and scaling to unit variance). The targets $u$ and $\mathbf i_\tau$ have been scaled to $[0,1]$ in order to match the output of the last sigmoid activations. During preliminary experiments, we found that for the value function network and the value function gradient network, a better performance was achieved by omitting any normalization of the data.



We split the data into a train, validation and test set by a ratio of 0.8/0.1/0.1. All networks were trained on mini-batches of size $b=8192$ using the Amsgrad optimizer with an initial learning rate of $10^{-5}$ for 300 epochs. For each experiment, we select the network with the lowest mean square error on the validation set for all further computations presented in this work.


\begin{figure}[tb]
    \centering
    \includegraphics[width=0.7\textwidth]{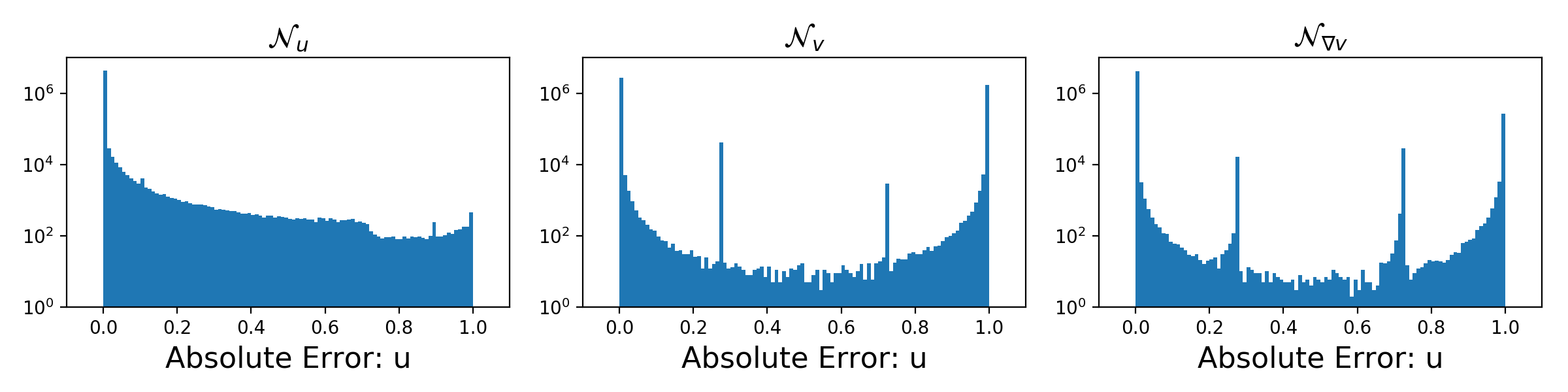}
    \caption{Distribution across the test set of the absolute error introduced when predicting the throttle using the three trained networks.
    \label{fig:throttledifference}}
\end{figure}

\begin{figure}[tb]
    \centering
    \includegraphics[width=1\textwidth]{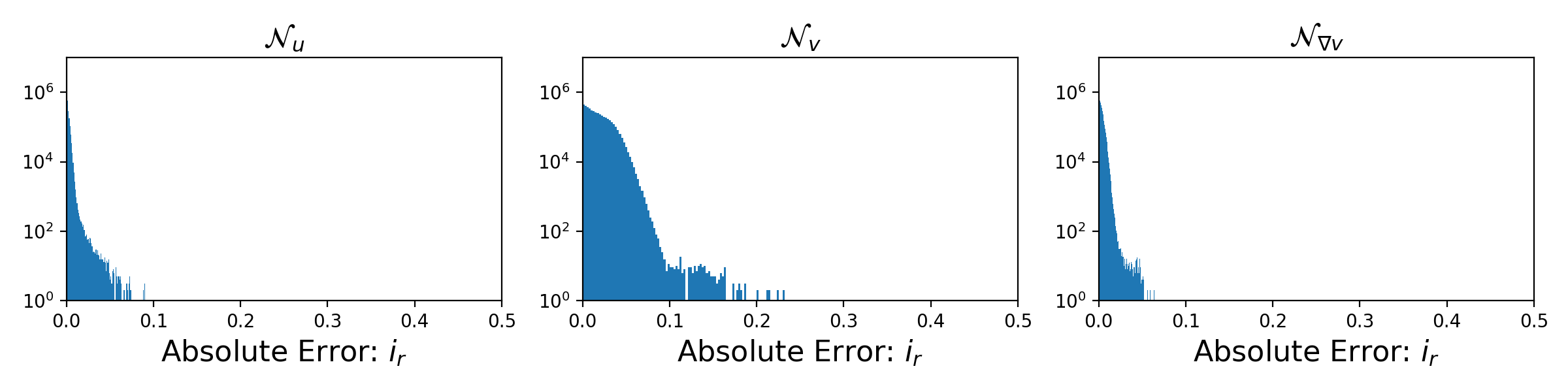} \\
    \includegraphics[width=1\textwidth]{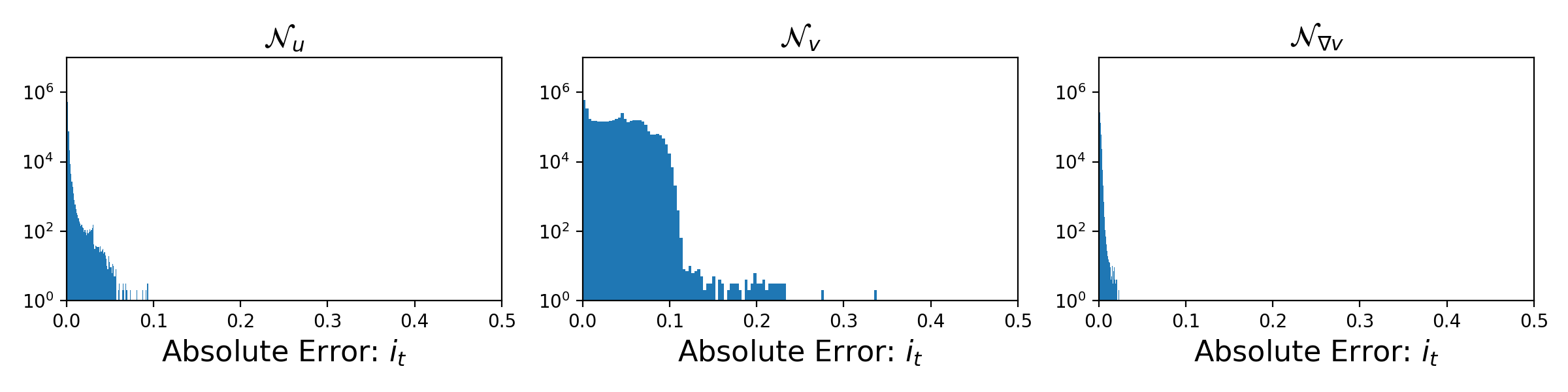} \\
    \includegraphics[width=1\textwidth]{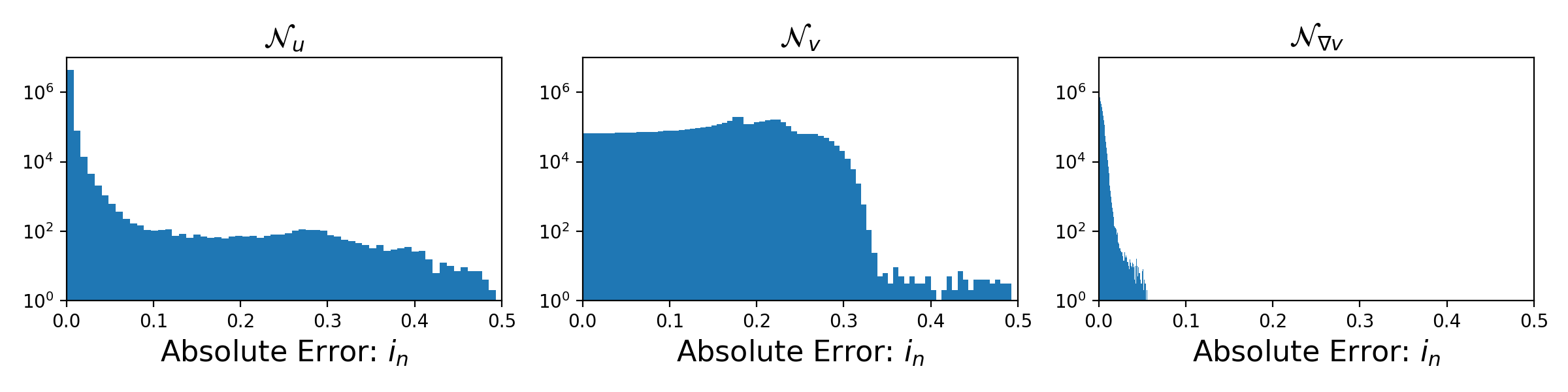}
    \caption{Distribution of thrust errors for the radial (top), tangential (middle) and normal (bottom) thrust components with a logarithmic y-scale.
    \label{fig:thrustdifference}}
\end{figure}

\section{Results}
\label{sec:results}

We evaluate the quality of the trained networks by two different methods. First, we test the network performance on the test set, containing optimal state control pairs from trajectories never seen during training. Second, and more importantly, we test how well the network performs when used to control the spacecraft dynamics, as computed by integrating the equations of motion in Eq.(\ref{eq:eom}) forward in time starting from the initial state of the nominal trajectory (which was excluded from the training, validation and test sets) using the policies resulting from the network for $\mathbf u$.

\subsection{Evaluation of Control Predictions on the Test Set}

    

\begin{table}[]
    \centering
    \begin{tabular}{c||c|c|c|c}
        controller & $u$ & $i_r$ & $i_t$ & $i_n$ \\ 
        \hline 
        $\mathcal{N}_{\mathbf u}$         & $4.1\cdot 10^{-3}\pm 1.6\cdot 10^{-3}$ & $9.4\cdot 10^{-4}\pm2.0\cdot 10^{-6}$ & $5.4\cdot 10^{-4}\pm8.8\cdot 10^{-7}$ & $2.1\cdot 10^{-3}\pm5.0\cdot 10^{-5}$ \\
        $\mathcal{N}_{v}$ & $3.8\cdot 10^{-1}\pm2.3\cdot 10^{-1}$  & $1.8\cdot 10^{-2}\pm1.8\cdot 10^{-4}$ & $3.7\cdot 10^{-2}\pm7.3\cdot 10^{-4}$ & $1.6\cdot 10^{-1}\pm5.6\cdot 10^{-3}$ \\
        $\mathcal{N}_{\nabla v}$  & $6.7\cdot 10^{-2}\pm6.1\cdot 10^{-2}$  & $3.0\cdot 10^{-3}\pm6.2\cdot 10^{-6}$ & $8.5\cdot 10^{-4}\pm7.3\cdot 10^{-7}$ & $2.6\cdot 10^{-3}\pm4.7\cdot 10^{-6}$ \\ 
    \end{tabular}    
    \caption{Mean and Standard Deviation of the Absolute Error of network predictions for the three network types.}
    \label{tab:controlmae}
\end{table}

For each sample of the test set we compute the optimal policy prediction from the networks ($u$ and $\mathbf{\hat i_\tau} $) and thus the mean and standard deviation of the absolute error. 
In Table~\ref{tab:controlmae}, the values for each of the control components are shown. The policy network displays the best performance in terms of the $u$, $i_r$ and $i_t$ components. 
On the other hand, the value function gradients network outperforms the policy network in terms of the $i_n$ component with regards to the spread in errors across the test set despite an overall higher mean absolute error. The value function network without gradient learning, while approximating the value function reasonably well, fails to reconstruct the optimal policy. This suggests that learning the gradients in addition to the value function is necessary for the reconstruction of an optimal control profile.

A more detailed view of the error distribution is shown in Figure~\ref{fig:throttledifference} for the throttle and Figure~\ref{fig:thrustdifference} for the thrust direction, confirming that the value function gradients network mis-throttles (i.e. throttling when it should not and not throttling when it should) on average more frequently than the policy network. It is interesting to note, again, that in terms of the distribution of errors in the thrust direction, the value function gradients network achieves a narrower distribution of errors centered around 0 for the $u_n$ component of the thrust.

\subsection{Closing the Gap to Venus Orbit}

\begin{table}[]
    \centering
    \begin{tabular}{c||c|c|c}
        controller & $m_f$ [kg] & $\Delta m$ [kg] & $m_f + \Delta m$ [kg]\\ 
        \hline 
        optimal & 210.4742 & -- & 210.4742 (+ 0.0000)\\ 
        $\mathcal{N}_{\mathbf u}$ & 210.3197 & 0.1858 & 210.5055 (+ 0.0313)\\
        $\mathcal{N}_{v}$         & 151.1732 & 129.7239 & 280.8971 (+ 80.7503) \\
        $\mathcal{N}_{\nabla v}$  & 205.4894 & 5.6123 & 211.1017 (+ 0.6275) \\ 
    \end{tabular}    
    \caption{Propellant mass spent to reach Venus orbit from the nominal initial condition. This mass consists of the propellant mass $m_f$ spent using the considered controller to steer the spacecraft for a time $t_f^*$ and the mass ($\Delta m$) needed for an additional (optimal) corrective maneuver able to match Venus orbit exactly.}
    \label{tab:gapclose}
\end{table}
 
\begin{figure}[tb]
    \centering
    \includegraphics[width=0.7\columnwidth]{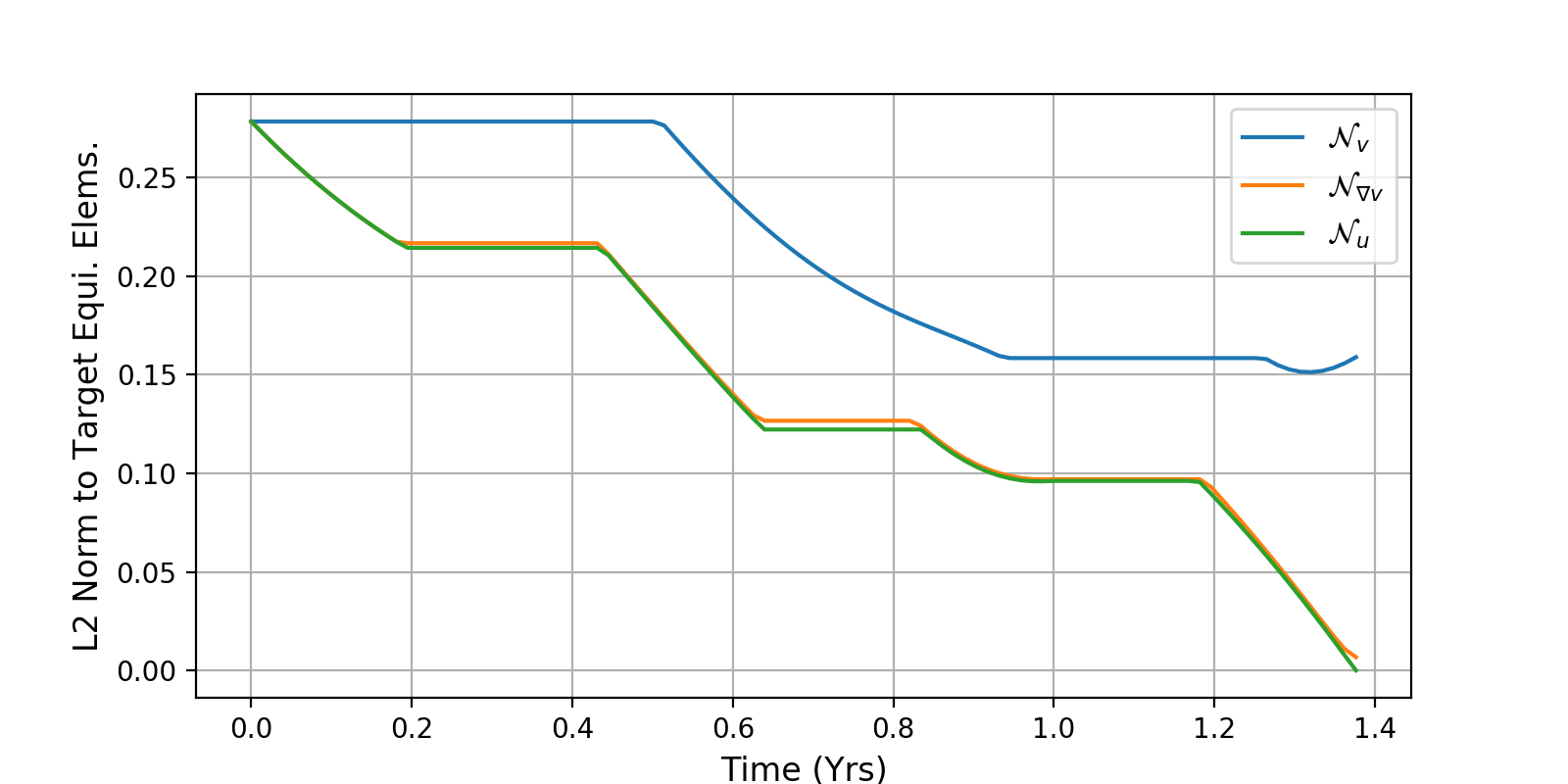}
    \caption{Distance from Venus during trajectory for value function (blue), value function gradients (orange) and policy networks (green).
    \label{fig:venusdist}}
\end{figure}

Next, we compare how well the trained neural networks are able to reproduce the nominal trajectory (which was not part of the dataset used for training) during a forward in time numerical integration of the equation of motion stated in Eq.(\ref{eq:eom}). Starting from the nominal initial conditions, we thus predict the control actions using each network and integrate forward in time for a period of $t_f^*= 1.376$ [years] (the time needed by the nominal trajectory to reach the orbit of Venus). Figure~\ref{fig:venusdist} shows the distance from the orbit of Venus obtained in the three cases. We observe that $\mathcal{N}_{\mathbf u}$ and $\mathcal{N}_{\nabla v}$ come closest to Venus orbit while $\mathcal{N}_{v}$ is comparatively far away.

For a fair comparison between all three strategies with respect to the nominal trajectory, we correct all orbits (in a mass-optimal way) to match the exact orbit of Venus and add the corresponding propellant mass ($\Delta m$) needed for this corrective maneuver to the propellant mass ($m_f$) spent by the controller. Consequently, we solve the optimal control problem for reaching Venus from each of the final orbits reached by the networks. The results are reported in Table~\ref{tab:gapclose}. We note that the additional mass needed by $\mathcal{N}_{\mathbf u}$ and $\mathcal{N}_{\nabla v}$ to close the gap to Venus orbit is below 1 [kg].

\section{Conclusions and Discussion}
\label{sec:conclusion}

The results presented show that our strategy for the generation of optimal trajectories is able to provide datasets of suitable size for supervised training of deep neural networks. The major benefit of this technique is the low computational cost and its reliability, as the usage of complex solvers for optimal control problems can be avoided in favor of a much simpler numerical integration procedure. For the test case of an interplanetary transfer to Venus orbit, we find that the small disturbances $\rho$, which we applied to the final co-states, allowed us to generate enough variety within the dataset to learn reliable control policies by imitation learning. While the policy network is structurally simple and comparatively small in size, the achieved precision is promising and provides a first step towards the goal of achieving on-board optimal guidance and control by a such a neural network.

Additionally to the prediction of the optimal control policy, the value networks are able to estimate the value function up to high precision and thus provide an approximation to the mass-budget needed for optimal transfers from different starting points. If the error of the gradients of the value function is incorporated during training, a control profile equally good as the imitation learning can be obtained.

While these preliminary results are encouraging, we plan to further study the potential of this approach for designing more ambitious transfers. In particular, the perturbation parameter of the final co-states $\rho$ is the key for this approach as it significantly impacts the resulting trajectory dataset, its needed size and, subsequently, the quality and usability of the trained networks. Limiting $\rho$ to a comparatively small value allowed the trajectories in our dataset to resemble the nominal trajectory closely. However, the investigation of larger values of $\rho$ is of great interest and remains for future work. The dataset size and its relation to $\rho$ is also of great interest, since in this work they were both selected in a somewhat arbitrary fashion.

Sampling of the trajectories at equally spaced points in time resulted in a bias of the dataset, as it leads to introduce more sample points at larger distances from the target orbit. Although this did not pose a problem during our experiments, a different sampling strategy could easily reduce the bias and would be most likely beneficial for training on challenging transfers. Furthermore, foregoing the computation of a nominal trajectory seems entirely possible and worth investigating, for example by generating the final co-states at random (and not perturbing some initial values). 

Lastly we note that for the value function gradient learning as presented here, the co-states need to be known (i.e. part of the dataset). While our approach is able to deliver the co-states for each trajectory, a direct method for solving the optimal control problem would not. In such cases we hypothesize that adding Eq.(\ref{eq:HJB1}) to the loss function might improve the quality of the controller even if information on the co-states is not available.

\appendix
\section{Appendix}
\label{app}
This appendix contains the explicit forms of all the derivatives necessary to write Eq.(\ref{eq:eom_costates}) explicitly.
\subsection*{The $\dot\lambda_p$ equation}
We get:

\begin{align}
\dot\lambda_p & = - \frac {c_1 u} m \boldsymbol\lambda^T \frac{\partial \mathbf B(\mathbf x)}{\partial p}  \mathbf i_\tau -  w^2 \lambda_L \frac{\partial}{\partial p} \sqrt{\frac{\mu}{p^3}} \\ &
=  - \frac {c_1 u} m \boldsymbol\lambda^T \frac{\partial \mathbf B(\mathbf x)}{\partial p}  \mathbf i_\tau  +\frac 32  w^2 \lambda_L \sqrt{\frac \mu{p^5}}  \nonumber
\end{align}
\begin{multline}
 2 \sqrt{\mu p} \frac{\partial \mathbf B(\mathbf x)}{\partial p} =  \\ =
\left[
\begin{array}{ccc}
\begin{array}{ccc}
0 &  \frac {6p}w  & 0 \\
 \sin L & [(1+w)\cos L + f]\frac 1w  & - \frac gw (h\sin L-k\cos L)  \\
- \cos L & [(1+w)\sin L + g]\frac 1w  & \frac fw (h\sin L-k\cos L)  \\
0 & 0  & \frac 1w \frac{s^2}{2}\cos L \\
0 & 0  & \frac 1w \frac{s^2}{2}\sin L \\
0 & 0  & \frac 1w (h\sin L - k\cos L) \\
\end{array}
\end{array}
\right]
\end{multline}

\subsection*{The $\dot\lambda_f$ equation}
We get:

\begin{align}
\dot\lambda_f & = - \frac {c_1 u} m \boldsymbol\lambda^T \frac{\partial \mathbf B(\mathbf x)}{\partial f}  \mathbf i_\tau - 2 \lambda_L w \sqrt{\frac{\mu}{p^3}}  \frac{\partial w}{\partial f}  \\ &
= - \frac {c_1 u} m \boldsymbol\lambda^T \frac{\partial \mathbf B(\mathbf x)}{\partial f}  \mathbf i_\tau - 2 \lambda_L w \sqrt{\frac{\mu}{p^3}} \cos L   \nonumber
\end{align}

\begin{multline}
 w^2 \sqrt{\frac \mu p} \frac{\partial \mathbf B(\mathbf x)}{\partial f} = \\ =
\left[
\begin{array}{ccc}
\begin{array}{ccc}
0 & - 2p \cos L  & 0 \\
0 & w - (\cos L + f)\cos L & g\cos L (h\sin L-k\cos L)    \\
0 & -(\sin L + g)\cos L &  (w-f\cos L) (h\sin L-k\cos L)  \\
0 & 0  & - \frac{s^2}{2}\cos^2 L \\
0 & 0  & -  \frac{s^2}{2}\sin L\cos L \\
0 & 0  & - (h\sin L - k\cos L) \cos L\\
\end{array}
\end{array}
\right]
\end{multline}

\subsection*{The $\dot\lambda_g$ equation}
We get:

\begin{align}
\dot\lambda_g & = - \frac {c_1 u} m \boldsymbol\lambda^T \frac{\partial \mathbf B(\mathbf x)}{\partial g}  \mathbf i_\tau - 2 \lambda_L w \sqrt{\frac{\mu}{p^3}}  \frac{\partial w}{\partial g}  \\ &
= - \frac {c_1 u} m \boldsymbol\lambda^T \frac{\partial \mathbf B(\mathbf x)}{\partial g}  \mathbf i_\tau - 2 \lambda_L w \sqrt{\frac{\mu}{p^3}} \sin L   \nonumber
\end{align}

\begin{multline}
 w^2 \sqrt{\frac \mu p} \frac{\partial \mathbf B(\mathbf x)}{\partial g} = \\ =
\left[
\begin{array}{ccc}
0 & - {2p} \sin L  & 0 \\
0 & - (\cos L + f) \sin L & - (w-g\sin L) (h\sin L-k\cos L)  \\
0 & w - (\sin L + g)\sin L & - f \sin L (h\sin L-k\cos L)   \\
0 & 0  & - \frac{s^2}{2}\cos L\sin L \\
0 & 0  & - \frac{s^2}{2}\sin^2 L \\
0 & 0  & -(h\sin L - k\cos L) \sin L\\
\end{array}
\right]
\end{multline}

\subsection*{The $\dot\lambda_h$ equation}
We get:

\begin{align}
\dot\lambda_h & = - \frac {c_1 u} m \boldsymbol\lambda^T \frac{\partial \mathbf B(\mathbf x)}{\partial h}  \mathbf i_\tau
\end{align}

$$
\frac{\partial \mathbf B(\mathbf x)}{\partial h} =  \sqrt{\frac p \mu}
\left[
\begin{array}{ccc}
0 & 0 & 0 \\
0 & 0  & - \frac gw \sin L  \\
0 & 0  & \frac fw \sin L  \\
0 & 0  & \frac hw \cos L \\
0 & 0  & \frac hw \sin L \\
0 & 0  & \frac 1w \sin L \\
\end{array}
\right]
$$


\subsection*{The $\dot\lambda_k$ equation}
We get:

\begin{align}
\dot\lambda_k & = - \frac {c_1 u} m \boldsymbol\lambda^T \frac{\partial \mathbf B(\mathbf x)}{\partial k}  \mathbf i_\tau
\end{align}

$$
\frac{\partial \mathbf B(\mathbf x)}{\partial k} =  \sqrt{\frac p \mu}
\left[
\begin{array}{ccc}
0 & 0  & 0 \\
0 & 0  & \frac gw\cos L  \\
0 & 0  & - \frac fw \cos L  \\
0 & 0  & \frac kw \cos L \\
0 & 0  & \frac kw \sin L \\
0 & 0  & -\frac 1w \cos L \\
\end{array}
\right]
$$

\subsection*{The $\dot \lambda_L$ equation}
We get:

\begin{equation}
\dot\lambda_L = - \frac {c_1 u} m \boldsymbol\lambda^T \frac{\partial \mathbf B(\mathbf x)}{\partial L}  \mathbf i_\tau - 2 w \sqrt{\frac{\mu}{p^3}} \lambda_L w_L
\end{equation}
\normalsize
where,

\begin{multline}
w^2 \sqrt{\frac \mu p} \frac{\partial \mathbf B(\mathbf x)}{\partial L} = \\
\left[
\begin{array}{ccc}
0 & -2p (g\cos L - f\sin L) & 0 \\
w^2\cos L & - (1+w)w\sin L - w_L(\cos L + f)  & ((wh+w_Lk)\cos L + (wk-w_Lh)\sin L) g  \\
w^2\sin L & (1+w)w\cos L - w_L(\sin L + g)  & ((wh+w_Lk)\cos L + (wk-w_Lh)\sin L) f  \\
0 & 0  & -\frac{s^2}{2}(w\sin L + w_L\cos L) \\
0 & 0  &  \frac{s^2}{2}(w\cos L - w_L\sin L) \\
0 & 0  &  {(wh+w_Lk)\cos L + (wk-w_Lh)\sin L} \\
\end{array}
\right]
\end{multline}

where,

\begin{equation}
w_L = \frac{\partial w}{\partial L} = g\cos L - f\sin L
\end{equation}
\normalsize

\subsection*{The $\dot \lambda_m$ equation}
We get:

\begin{align}
\dot\lambda_m & = - \frac {c_1 u} {m^2} |\boldsymbol\lambda^T \mathbf B(\mathbf x)|
\end{align}

\bibliographystyle{acm}
\bibliography{arxiv} 

\begin{thebibliography}{10}

\bibitem{bellman1966dynamic}
{\sc Bellman, R.}
\newblock Dynamic programming.
\newblock {\em Science 153}, 3731 (1966), 34--37.

\bibitem{bertrand2002new}
{\sc Bertrand, R., and Epenoy, R.}
\newblock New smoothing techniques for solving bang--bang optimal control
  problems--numerical results and statistical interpretation.
\newblock {\em Optimal Control Applications and Methods 23}, 4 (2002),
  171--197.

\bibitem{cheng2019real}
{\sc Cheng, L., Wang, Z., Song, Y., and Jiang, F.}
\newblock Real-time optimal control for irregular asteroid landings using deep
  neural networks.
\newblock {\em arXiv preprint arXiv:1901.02210\/} (2019).

\bibitem{furfaro2018deep}
{\sc Furfaro, R., Bloise, I., Orlandelli, M., and Di~Lizia, P.}
\newblock Deep learning for autonomous lunar landing.
\newblock In {\em 2018 AAS/AIAA Astrodynamics Specialist Conference\/} (2018),
  pp.~1--22.

\bibitem{haberkorn2004low}
{\sc Haberkorn, T., Martinon, P., and Gergaud, J.}
\newblock Low thrust minimum-fuel orbital transfer: a homotopic approach.
\newblock {\em Journal of Guidance, Control, and Dynamics 27}, 6 (2004),
  1046--1060.

\bibitem{izzo2018survey}
{\sc Izzo, D., M{\"a}rtens, M., and Pan, B.}
\newblock A survey on artificial intelligence trends in spacecraft guidance
  dynamics and control.
\newblock {\em arXiv preprint arXiv:1812.02948\/} (2018).

\bibitem{izzo2018machine}
{\sc Izzo, D., Sprague, C., and Tailor, D.}
\newblock Machine learning and evolutionary techniques in interplanetary
  trajectory design.
\newblock {\em arXiv preprint arXiv:1802.00180\/} (2018).

\bibitem{pontryagin}
{\sc Pontryagin, L., Boltyanskii, V., Gamkrelidze, R., and Mishchenko, E.}
\newblock {\em The Mathematical theory of optimal processes}.
\newblock Wiley \& Sons, 1962.

\bibitem{sanchez2018real}
{\sc S{\'a}nchez-S{\'a}nchez, C., and Izzo, D.}
\newblock Real-time optimal control via deep neural networks: study on landing
  problems.
\newblock {\em Journal of Guidance, Control, and Dynamics 41}, 5 (2018),
  1122--1135.

\bibitem{sanchez2016learning}
{\sc S{\'a}nchez-S{\'a}nchez, C., Izzo, D., and Hennes, D.}
\newblock Learning the optimal state-feedback using deep networks.
\newblock In {\em 2016 IEEE Symposium Series on Computational Intelligence
  (SSCI)\/} (2016), IEEE, pp.~1--8.

\bibitem{sprague2019learning}
{\sc Sprague, C.~I., Izzo, D., and {\"O}gren, P.}
\newblock Learning a family of optimal state feedback controllers.
\newblock {\em arXiv preprint arXiv:1902.10139\/} (2019).

\bibitem{tailor2019learning}
{\sc Tailor, D., and Izzo, D.}
\newblock Learning the optimal state-feedback via supervised imitation
  learning.
\newblock {\em arXiv preprint arXiv:1901.02369\/} (2019).

\bibitem{walker}
{\sc {Walker}, M.~J.~H., {Ireland}, B., and {Owens}, J.}
\newblock {A set of modified equinoctial orbit elements}.
\newblock {\em Celestial Mechanics 36\/} (Aug. 1985), 409--419.

\bibitem{zhu2019fast}
{\sc Zhu, Y.-h., and Luo, Y.-z.}
\newblock Fast approximation of optimal perturbed many-revolution
  multiple-impulse transfers via deep neural networks.
\newblock {\em arXiv preprint arXiv:1902.03741\/} (2019).

\end{thebibliography}

\end{document}